\documentclass[conference]{IEEEtran}

\usepackage{graphicx}
\usepackage{amsmath} 
\usepackage{float}
\usepackage{url}
\usepackage{tabu}
\usepackage{algorithm2e}
\usepackage{balance}

\begin{document}

\title{
    Rapid-Rate: A Framework for Semi-supervised\\Real-time Sentiment Trend Detection in\\Unstructured Big Data\\
    \vspace{5mm}
    \large CS 846 (Fall 2016) Course Project
}

\author{
    \IEEEauthorblockN{Vineet John}
    \IEEEauthorblockA{
        David R. Cheriton School of Computer Science\\
        University of Waterloo\\
        Waterloo, Ontario N2L 3G1\\
        Email: vineet.john@uwaterloo.ca
    }
}

\maketitle

\begin{abstract}
    Commercial establishments like restaurants, service centers and retailers have several sources of customer feedback about products and services, most of which need not be as structured as rated reviews provided by services like Yelp, or Amazon, in terms of sentiment conveyed.
    For instance, Amazon provides a fine-grained score on a numeric scale for product reviews.
    Some sources, however, like social media (Twitter, Facebook), mailing lists (Google Groups) and forums (Quora) contain text data that is much more voluminous, but unstructured and unlabeled.
    It might be in the best interests of a business establishment to assess the general sentiment towards their brand on these platforms as well. 
    
    This text could be pipelined into a system with a built-in prediction model, with the objective of generating real-time graphs on opinion and sentiment trends.
    Although such tasks like the one described about have been explored with respect to document classification problems in the past, the implementation described in this paper, by virtue of learning a continuous function rather than a discrete one, offers a lot more depth of insight as compared to document classification approaches. 
    
    This study aims to explore the validity of such a continuous function predicting model to quantify sentiment about an entity, without the additional overhead of manual labeling, and computational preprocessing \& feature extraction.
    This research project also aims to design and implement a re-usable document regression pipeline as a framework, Rapid-Rate\cite{rapid_rate}, that can be used to predict document scores in real-time.
\end{abstract}

\vspace{5mm}

\providecommand{\keywords}[1]{\textbf{\textit{Keywords---}} #1}

\IEEEpeerreviewmaketitle

\vspace{5mm}

\section{Introduction}
    Document labeling and significant feature discovery are already widely researched areas in the domain of natural language processing.
    The most common use-cases of document classification are product \& service review rating predictions, automatic grading of essays, spam detection, plagiarism detection etc.
    However, most of the current approaches used for these tasks rely on lexicon based methods with a centralized dictionary indicating the weights of a word towards a set of emotions or sentiment.
    Some other approaches use document level similarity scores like TF-IDF or n-gram count vectorization approaches to assess the class of a previously unseen document.

    However, models like these are difficult to train for the following reasons:
    \begin{itemize}
      \item Having to hand-pick features
      \item Having to use stepwise regression to eliminate non-relevant features
      \item Having to manually re-train the models on a subset of the features to determine the best-fit
    \end{itemize}

    All of the processes described above are time-consuming and tedious.
    Also, n-gram approaches only partially model the language context probabilities i.e. the probably of a word occurring next in the corpus depends only on the n-1 words that precede it, and TF-IDF loses word context altogether.
    The objective of this research project is to implement and evaluate a software framework to effectively eliminate the above steps from the process of training a language model. 
    Word2Vec, for instance, takes additional word context features into account while building the language model, a feature which will be elaborated upon in Section \ref{Word2Vec}.

    The proposition is to rely on unsupervised strategies like continuous bag-of-words and the skip-gram model to learn word embeddings.
    These word embeddings will serve as a syntactic and semantic proxy for the documents being processed. 
    The features thus learned, can then be used to train regression models to predict document scores.

\begin{figure*}[ht]
    \centering
    \includegraphics[width=\textwidth]{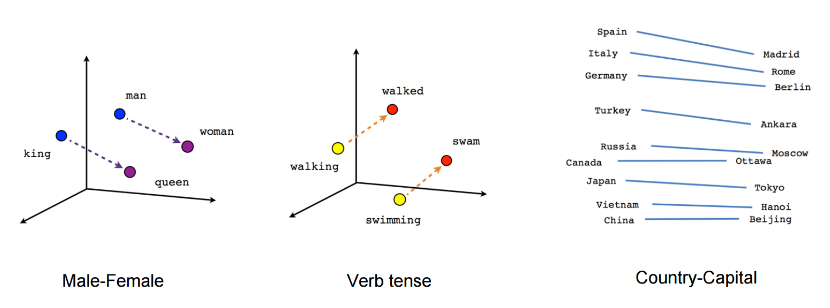}
    \caption{Word2Vec Vector Space Intuition\cite{tensorflow_word2vec}}
    \label{fig:word2vec-vectorspace-intuition}
\end{figure*}

\vspace{5mm}

\section{Similar Work}
    This section discusses the previous work done on mining and analytics pertaining to text data.

    Most of the similar work in the area of learning word-embeddings is related to classifying documents into a given set of labels or topics. 
    The purpose of this study is to extend the usage of the document vector representation to demonstrating its usage when the response variable (the output of a learning task) a continuous function (document-scoring) rather than discrete values (document classification). 
    This work can be viewed as a generalization of the previous approaches, because this study aims to predict fine-grained document scores, rather than coarse document classifications.

    Most of the current work in usage of regression techniques on text content are related to meta-data or features extraction \cite{su2015genetic}\cite{weissman2016natural}.
    Similarly there have been several studies to identify document tags using classification techniques\cite{bespalov2011sentiment}\cite{pang2002thumbs}. 
    However, there is a relative absence of studies to evaluate the accuracy of regression models that try to predict a document score using a continuous-function model.

    As far the the author is aware, there has been no similar work to test the regression accuracy of paragraph vectors in an experimental research setup before, and is hence, novel in that respect.
    The evaluation of the effectiveness of this approach will be assessed using a generic evaluation metric, the coefficient of determination score, which compares the predicted score-set against the golden set of actual scores.

\vspace{5mm}

\section{Problem Statement} \label{problem_statment}
    The problem statement can be formulated as an evaluation of Paragraph Vectors\cite{le2014distributed} on a document score regression task to evaluate the accuracy of the prediction model that can be built using the shallow neural network language model learning algorithm previously described by Mikolov et al.\cite{mikolov2013efficient}, which was primarily used to train word embeddings in an unsupervised manner.

    The prediction scores will be evaluated using the co-efficient of determination metric, also known as the R\textsuperscript{2} metric\cite{cameron1997r}. 
    The hypothesis is to prove that the document ratings have a positive correlation with the vectorized versions of the document text content, because the semantics of positive and negative reviews are expected to be captured by the shallow neural network trained on the corpus.
    The expected R\textsuperscript{2} metric score being over 0, will confirm the hypothesis, and a negative value will contradict it.

\begin{figure*}[ht]
    \centering
    \includegraphics[width=400pt]{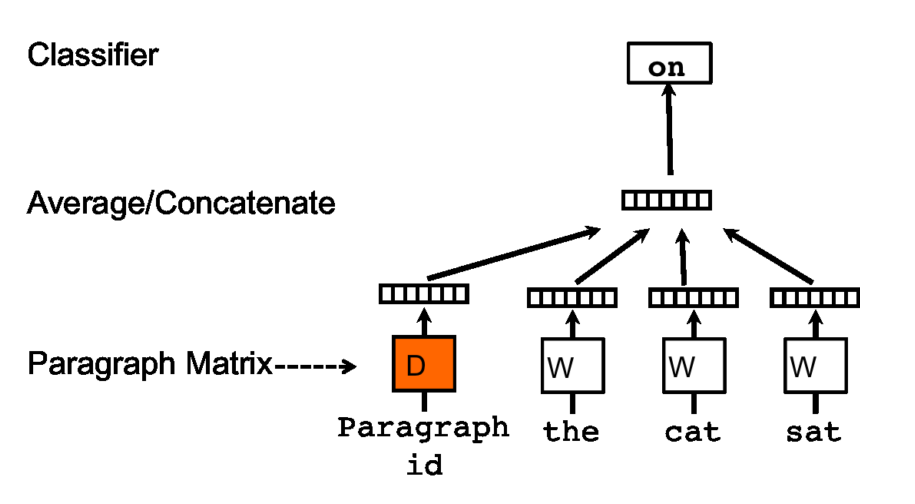}
    \caption{Paragraph Vector Learning Framework\cite{mikolov2013distributed}}
    \label{fig:paragraph-vector-framework}
\end{figure*}

\vspace{5mm}

\section{Motivation}
    The motivation for the project is to prove that a substantial amount of the semantic signal about sentiment with respect to a product or service, as extracted from a snippet or document of text, should prove to be a good indicator of the labeled document scores. 
    If proven to be true, the objective includes building a real-time streaming framework, which can analyze unseen reviews from the corpus of unlabeled data, and be able to emit a score-prediction with minimal processing delays.
    This framework can be set-up on commodity hardware, and a given business can simply pipe their unlabeled reviews or articles into the system, and expect a graphical representation of the sentiment about a particular topic that is updated in real-time. 
    The framework also allows rules to be set for alerts, so, for instance, if the approval rating of a particular business falls below a user-defined threshold, an email alert could be triggered to the business decision makers to respond quicker to the negative sentiment, a process which would otherwise span several business days.

    \subsection{Semi-supervised Learning}
        The statistical learning formulation of the problem statement in this study, can be broken down into two distinct segments, namely:
        \begin{itemize}
            \item Learning a projection of the document into vector space. This projection will act as a vector-space proxy for the sentiment that the document is trying to convey. This segment is completely unsupervised, and does not require any tuning or preprocessing to compute.
            \item Learning a relation between the aforementioned projection of each labeled document, and the associated score or label. This segment of the learning pipeline is supervised relies on previously scored or rated documents.
        \end{itemize}

    \subsection{Continuous function learning}
        Assume that a given output variable \textit{y} depends on a variable \textit{x} such that
        \begin{equation}
            \displaystyle f(x) = \beta x + c + \epsilon
        \end{equation}

        \begin{equation}
            \displaystyle y = f(x)
        \end{equation}
        where x represents a vector of independent variables, y represents the dependent or response variable, $\beta$ and c are model vector space functions that are applied to x to map it to y, and $\epsilon$ is the error term.

        Now, assume that the variable \textit{x} increases by a small amount $\Delta x$, and that $x_0$ was the initial value, such that the value of \textit{x} is updated to
        \begin{equation}
            \displaystyle \Delta x = x - x_0
        \end{equation}

        A continuous function is defined as one in which an infinitely small change for the input variable x results in a corresponding infinitely small change in the response variable y\cite{continuous_function}, which can be expressed as
        \begin{equation}
            \displaystyle \Delta y = f(x) - f(x_0)
        \end{equation}

        Continuous functions are typically needed when the response variable calculated is a measurable quantity, rather than a class label. Since sentiment as fine-grained values offer more depth in terms of sentiment insight \cite{drake2008sentiment}, the study in this paper is formulated as a regression problem.

\vspace{5mm}

\section{Natural Language Understanding Domain Challenges}
    The section elaborates on the NLP domain-specific challenges faced during the implementation of this research project, and the proposed strategies used to tackle these challenges or minimize their impact on the final outcome of the experiment.

    \subsection{Language Model Learning}
        Language model learning is complicated because the most popular modeling approaches currently used have inherent limitations when it comes to being able to be used reliably for language understanding accuracy across the board. A few such instances are described in this section.
        
        \subsubsection{Lexicon based model learning}
            Lexicon based models are reliable for one-off tasks, but they require hand-annotation of the polarity of words, and an over-dependence on the fact that the lexicon of words needs to be updated regularly as and when it comes across new words. 
            This requires time and effort and is rather tedious. 
            This challenge is addressed in the research project, by using an unsupervised method to learn the document vector representations, and not relying on any lexicon for details about word or document semantics.

        \subsubsection{Statistical model based learning}
            TF-IDF and similar document scoring approaches, are based on the bag-of-words model. Hence, it does not capture position in text, semantics, co-occurrences in different documents, and other such features. Also, this shortcoming means that it cannot capture semantics of a word. It relies heavily on occurrence count across the entire corpus of documents.

        \subsubsection{Cross domain applicability} \label{cross_domain_applicability}
            One of the major drawbacks of statistical model based learning in particular is that a language model trained on a dataset belonging to a particular domain is highly unlikely to yield good results when assessed or tested on another domain. 
            For instance, a language model trained for detecting sentiment on a movie review data-set can't be used to accurately predict sentiment on a stock-market discussion dataset. 
            This is due to the inherent business domain specific differences in the vocabulary and connotations that will unlikely be modeled when trained on a completely different dataset than the one that the model needs to be tested upon. 
            This problem is not directly tackled for this study, but the author proposes a potential direction in the Section \ref{future_work} which may help address this problem in future versions of Rapid-Rate.

    \subsection{Sarcasm/Irony Detection in Text}
        One of the biggest hurdles with accurately inferring sentiment of a document despite having a robust trained model, or even a lexicon, is the presence of sarcasm and irony, which are extremely likely for unmoderated comments and articles, especially on forums and social media. 
        For now, there is no clear best-approach for sarcasm and irony detection, and often, the best features for detecting such sentiment is text meta-data like emoticons and tags rather than the text itself, as discussed by Gonz{\'a}lez-Ib{\'a}nez et al.\cite{gonzalez2011identifying}.

    \subsection{Deep Learning}
        The challenge with using deep learning for the model training is there deep learning models typically take hours to even days to completely train. 
        If a system needs to be setup from scratch, deep learning does increase the time to market by being notoriously difficult to train. 
        This challenge addressed by incurring only a one-time cost of training the labeled document corpus. Future iterations of the training does not block the current execution of the real-time prediction component.

\vspace{5mm}

\section{Engineering Design \& Implementation Challenges}
    The previous section described the natural language understanding challenges. This sections aims to describe the engineering design and implementation challenges that needed to be tackled during the implementation phase.

    \subsection{Unstructured Big Data}
        Unstructured data is defined as data, large amounts of it, in this case, which do not abide by a strict schema\cite{buneman1996query}. 
        This happens when the process of data generation itself is unmoderated and does not impose too many constraints on the type of data allowed. 
        It could also be that the data contains soft links or references to other entities within the data. 
        As such, the onus is on the application to process large amount of unstructured data and attempt to extract some meaning and relations from it.
        This is tackled by adopting a Paragraph Vector approach to convert variable-length text into fixed-length vectors, described in detail in Section \ref{paragraph_vectors}.

    \subsection{Distributed Document Vector Learning} \label{dist_docvec_learning}
        The preference for the most computationally intensive tasks is to distribute them by scaling horizontally.
        This allows a driver to submit multiple chunks of the same job to a number of workers.
        These workers can then convey the output to the driver once they finish their respective jobs, and the driver can aggregate these results.
        However, the library used for this study, Gensim, does not, by default, include a distributed setup for document model learning.
        Instead the onus is on the implementation to integrate Gensim with a distributed processing framework like Spark or Hadoop.
        The current implementation of Rapid-Rate uses a low-memory footprint iterator based technique to reduce the system memory usage on a single node, for the off-line document vector learning task described in Section \ref{offline_module}.
        This could be scaled to train even higher volumes of documents if integrated with a distributed processing framework.

    \subsection{Optimal Setup for Production Execution}
        The motivation behind building a framework, is to be able to quickly deploy it to a typical production environment in an industrial setting.
        Hence the decision to separate the business logic of the document vector training component (off-line module) and the real-time prediction component (on-line module), as modular set.
        This can enable the document vector training to be done asynchronously in the background, agnostic of the current state of the real-time component, and keep enhancing the currently learned document vector model and machine learning modules.
        Similarly, the real-time prediction component does not need to be blocked on the document vector learning component after the first run.
        It can simply keep using the older version of the trained document vector model and machine learning model while the off-line component's scheduled cronjob runs in the background.

        This allows for easy maintenance if the off-line module needs to be temporarily paused for maintenance.
        Similarly, if the always-running daemon job for the on-line module needs to be temporarily paused, the streaming documents are not lost. 
        The distributed queue keeps the documents queued, along with the offset of where the on-line module needs to resume reading from. 
        This ensures that the on-line module can simply proceed from where it left off before being taken down for maintenance.

\vspace{5mm}

\begin{figure*}[ht]
    \centering
    \includegraphics[width=\textwidth]{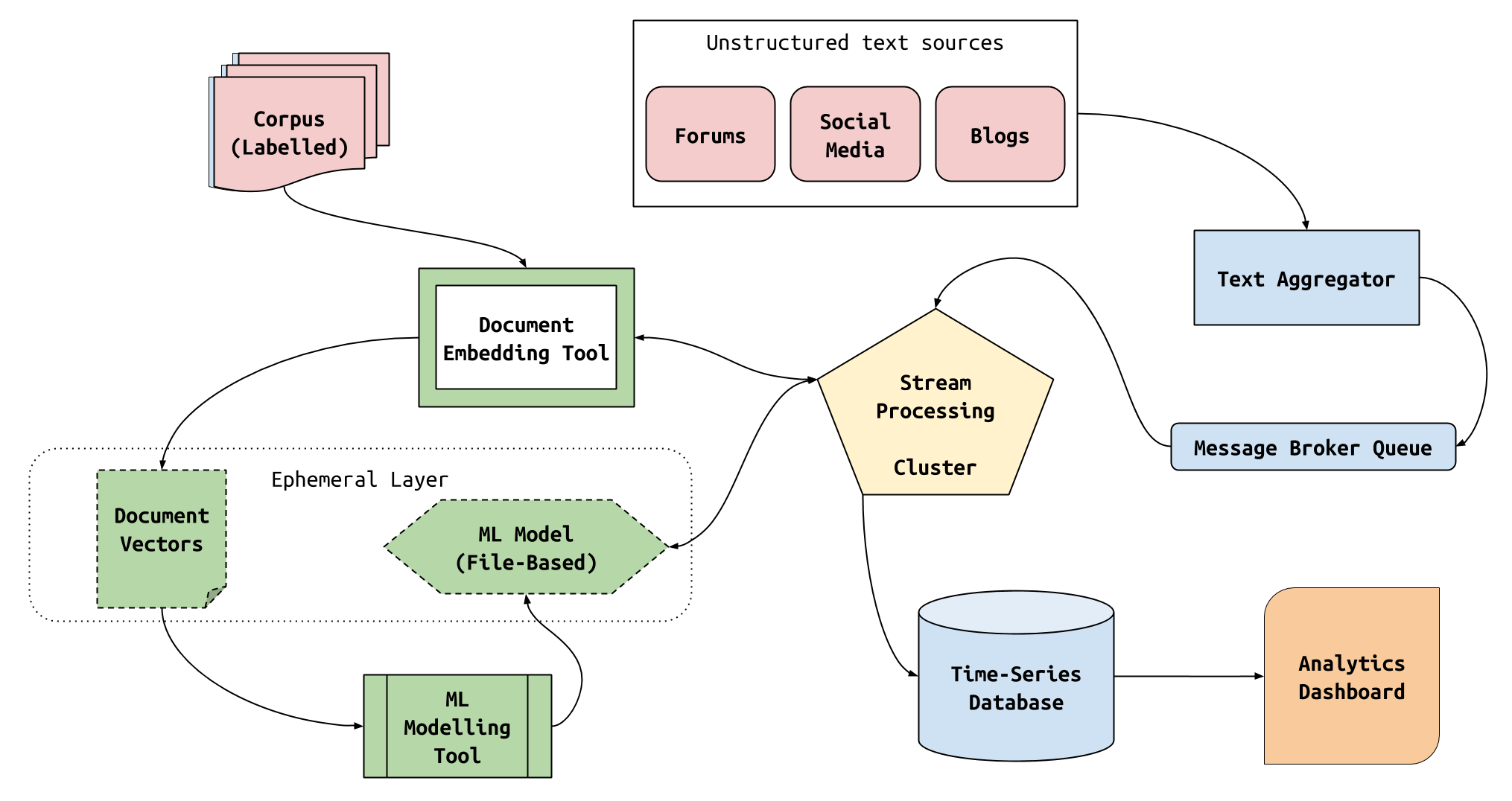}
    \caption{Rapid-Rate System Architecture}
    \label{fig:system-architecture}
\end{figure*}

\section{Background}
    The sections touches upon the background information that needs to be known prior to doing a deep-dive into the details of the research study and implementation done for this project.

    \subsection{Word2Vec} \label{Word2Vec}
        Word vectors are a step-away from the traditional bag-of-words + n-gram model. 
        The latter ends up losing context of the words because of the fact that the words that follow it are not taken into account, and only the preceding words are.

        The Word2Vec model\cite{mikolov2013efficient} can take into consideration both preceding and following words, as a sliding window of text, to compute the probability of occurrence of the current word, as part of the Continuous Bag-of-words (CBOW) model. 
        Alternatively, Word2Vec also provides a skip-gram model\cite{mikolov2013distributed}, which conversely, uses the context probabilitie of the current word, to predict the most likely words to precede and follow it. 
        These word embeddings can be used to derive both syntactic as well as semantic relations in the vector space of the language model. A representation of this is given in Figure \ref{fig:word2vec-vectorspace-intuition}.

    \subsection{Paragraph Vectors} \label{paragraph_vectors}
        Paragraph vectors are a similar concept to Word2Vec, but rely on computing a fixed-length vector representation of a variable length group of words, which could be a sentence, a paragraph, or even an entire document.

        Since vectors representations of documents seem to outperform bag-of-words models for the purposes of semantic summarization and document classification tasks, it follows that a vector representation based approach should be used to evaluate performance on a regression task as well. 
        Paragraph vectors also address some of the key weaknesses of bag-of-words models. 
        First, they inherit an important property of the word vectors: the semantics of the words. 
        The second advantage of the paragraph vectors is that they take into consideration the word order, at least in a small context, in the same way that an n-gram model with a large n would do. 
        This is important, because the n-gram model preserves a lot of information of the paragraph, including the word order\cite{le2014distributed}.
        The paragraph representation framework is depicted in Fig. \ref{fig:paragraph-vector-framework}. 

        The core concept here, is that an aggregation of the word vectors are being associated with a paragraph ID that encapsulates all of the words in question. 
        Thus, rather than just assigning each individual word a vector representation, each paragraph/group of words in the body of the text document have been assigned a vector representation.

        The implementation of paragraph vectors used for the purpose of implementing the system architecture in this study is Doc2Vec, from the Gensim library\cite{doc2vec_api}.

    \subsection{Statistical Learning Models}
        Once the entities about which we wish to learn a representation are projected into vector space, there are an abundance of statistical learning methods that can be used to learn a continuous function including linear regression, polynomial regression, ridge regression, stepwise regression, lasso regression et. al.

        This study utilizes linear regression and support vector regression to test the veracity of the model predictions.
        Both these methods are described in the below sub-sections.

        \subsubsection{Linear Regression}
            The problem of regression in mathematical statistics is characterized by the fact that there is insufficient information about the distributions of the variables under consideration\cite{regression_analysis}.

            In the case for this study, once the Doc2Vec implementation generates the vector space representation of each of the documents, each of the learned dimensions represent a different input variable for the regression problem. In reality, each of these dimensions represent a weighted probability score generated by the hidden layer of neurons in the paragraph vector model.

        \subsubsection{Support Vector Regression}
            Support Vector Regression (SVR) attempts to minimize the generalization error bound so as to achieve generalized performance. The idea of SVR is based on the computation of a linear regression function in a high dimensional feature space where the input data are mapped via a nonlinear function. SVR has been applied in various fields – time series and financial (noisy and risky) prediction, approximation of complex engineering analyses, convex quadratic programming and choices of loss functions\cite{basak2007support}.

            The input parameters to the support vector regression model remain the same as those described above for the linear regression model.

    \subsection{Distributed Messaging with Kafka}
        This project uses Apache Kafka for processing huge volumes of data from real-time streams.
        Like a messaging system, Kafka employs a pull-based consumption model that allows an application to consume data at its own rate and rewind the consumption whenever needed. 
        By focusing on log processing applications, Kafka achieves much higher throughput than conventional messaging systems. 
        It also provides integrated distributed support and can scale out. 
        Kafka has been used successfully at LinkedIn for both off-line and on-line applications.\cite{kreps2011kafka}

    \subsection{Horizontal Scaling with Spark Streaming}
        This study used Apache Spark Streaming for the real-time processing of the unlabeled document data. 
        Discretized streams (D-Streams), a stream programming model for large clusters that provides consistency, efficient fault recovery, and powerful integration with batch systems is the core offering of Spark Streaming. 
        The key idea is to treat streaming as a series of short batch jobs, and bring down the latency of these jobs as much as possible. 
        This brings many of the benefits of batch processing models to stream processing, including clear consistency semantics and a new parallel recovery technique that is a truly cost-efficient recovery technique for stream processing in large clusters\cite{zaharia2012discretized}.

    \begin{algorithm}[ht]
        \KwData{$document\_corpus$}
        \KwResult{$document\_vector\_model$, $stat\_learning\_model$}
        \DontPrintSemicolon 
        \;
        initialize $training\_passes$\;
        initialize $file\_iterator$\;
        initialize $docvec\_model$\;
        initialize $ml\_model$\;
        \;
        \For{$line \in file\_iterator$}{
            $parsed\_line = parse\_review(line)$\;
            $
            add\_to\_docvec\_model\_corpus($\\
            $\qquad parsed\_line.document\_text$\\
            $)$\;
            \For{$count \in training\_passes$}{
                $docvec\_model.train()$
            }
        }
        \;
        \For{$line \in file\_iterator$}{
            $parsed\_line = parse\_review(line)$\;
            $add\_to\_document\_scores($\\
            $\qquad parsed\_line.document\_score$\\
            $)$\;
        }
        \;
        \For{$i \in documents.count$}{
            $x\_docvecs[i] = docvec\_model.get\_vector(i)$\;
            $y\_scores[i] = document\_scores[i]$\;
        }
        \;
        $ml\_model = model.learn(x\_docvecs, y\_scores)$\;
        $ml\_model.predict\_accuracy\_score(y\_test, y\_true)$\;
        \;
        $docvec\_model.save\_to\_disk()$\;
        $ml\_model.save\_to\_disk()$\;
        \;
        \caption{Off-line learning algorithm}
         \label{offline_module_algo}
    \end{algorithm}

\vspace{5mm}

\section{System Architecture}
    This section describes the proposed architecture design, including the constituent components and the flow of the control and data throughout the architecture.
    The system architecture is depicted in Fig. \ref{fig:system-architecture}

    The individual components and their interactions are described in the below sub-sections \ref{Components} and \ref{Component Interactions}.

    \subsection{Components} \label{Components}

        \subsubsection{Document embedding tool}
            The labeled documents are the initial input parameters to the Rapid-Rate framework. 
            These are variable length text snippets or paragraphs, each with an associated score. 
            This tool is used to convert the document text into their corresponding vector representations.

        \subsubsection{ML Modeling tool}
            This is a library used by the framework to infer a model from the vectorized documents (independent variable) that predict the rating/score of that particular document (dependent variable). 
            It uses algorithms such as stochastic gradient descent in an attempt to converge to a lowest possible value of learning error\cite{bottou2010large}. 

        \subsubsection{Text Aggregation}
            This component is responsible for merging multiple different streams of data into a unified stream that can be used to predict sentiment ratings of unlabeled documents. 
            This can be implemented in frameworks like Apache Flume or Apache Storm, which allow definitions of multiple sources and sinks, with a topology for data processing as part of the aggregation business logic.

        \subsubsection{Message Broker Queue} \label{message_broker_queue}
            This component acts as the arbiter for the incoming unlabeled documents. 
            Each document is read from the queue and is processed in turn.
            The queue is capable of holding a large buffer, as well as an offset for each of it's consumers, to serve as a memory or bookmark of where the consumer stopped during their last subscription to the queue's updates.

        \subsubsection{Stream Processor}
            The stream processor is responsible for reading and processing the unlabeled documents, predicting the score/rating of each incoming document, and forwarding the scores for further usage.
            This component's setup is distributed, and is also able to subscribe to the updates being pushed by the message broker queue (Section \ref{message_broker_queue}). 

        \subsubsection{Time-Series Database}
            A time series database is optimized to handled collections of data objects that are indexed by time. 
            It supports fast querying of data over temporal scan ranges and is ideal for time-series based analytics

        \subsubsection{Analytics Dashboard}
            The analytics dashboard provides a visual representation of overall sentiment over time, and supports setting alerts to warn the end-user when overall sentiment dips below a certain user-defined threshold.

    \subsection{Component Interactions} \label{Component Interactions}
        A simple processor reads the labeled documents and feeds it into the Document embedding tool, which, in turn, emits a representation of each of the labeled documents as a vector.
        The vectors representations of the documents are now associated with their corresponding scores and the vector representations and the scores fed together into the ML Modeling tool to generate Linear Regression and Support Vector Regression models. 
        These models are then `pickled' or flushed to disk, to be used by the real-time prediction module. 
        Along with these models, the trained Document embedding model is also flushed to disk.

        Once these models are flushed to disk, the system can kick-start it's real-time prediction framework. 
        The unlabeled documents are published to the message broker queue, to which the daemon component of the architecture is subscribed to. 
        It processes lists of documents in micro-batches and emits a list of scores. 
        These scores are posted in real-time to a time-series database, which in turn, acts as the data back-end for the analytical dashboard.

        The analytical dashboard provides a line graph of the sentiment trend over time, and can also provided for additional features such as Bollinger Band plotting, and threshold breaching alerts.

\vspace{5mm}

\begin{algorithm}[ht] \label{online_module}
    \KwData{$document\_vector\_model$, $stat\_learning\_model$}
    \KwResult{$distributed\_queue\_uri$}
    \DontPrintSemicolon 
    \;
    initialize $kafka\_queue\_instance$\;
    initialize $spark\_context$\;
    initialize $spark\_streaming\_context$\;
    \;
    initialize $docvec\_model$\;
    initialize $ml\_model$\;
    initialize $kafka\_stream$\;
    \;
    \uIf{$new\_documents \in kafka\_queue\_instance$}{
        initialize $messages\_rdd = new\_documents$\;
        \For{$rdd \in messages\_rdd$}{
            $rdd.infer\_sentiment(docvec\_model, ml\_model)$
        }
    }
    \;
    \caption{Off-line learning algorithm}
\end{algorithm}

\begin{algorithm}[hb] \label{infer_sentiment}
    \KwData{$rdd$, $docvec\_model$, $ml\_model$}
    \KwResult{$null$}
    \DontPrintSemicolon 
    initialize $opentsdb\_connection$
    \;
    \For{$(key, message) \in rdd.collect()$}{
        $tokenized\_document = message.tokenize()$\;
        $document\_vector = docvec\_model.infer\_vector(tokenized\_document)$\;
        $predicted\_score = ml\_model.predict(document\_vector)$\;
        $tsdb\_payload = generate\_tsdb\_payload($\\ \qquad $predicted\_score, time.now()$\\$)$\;
        $tsdb\_payload.post()$
    }
    \;
    \caption{Sentiment Inference algorithm}
\end{algorithm}

\begin{figure*}[ht] 
    \centering
    \includegraphics[width=0.7\textwidth]{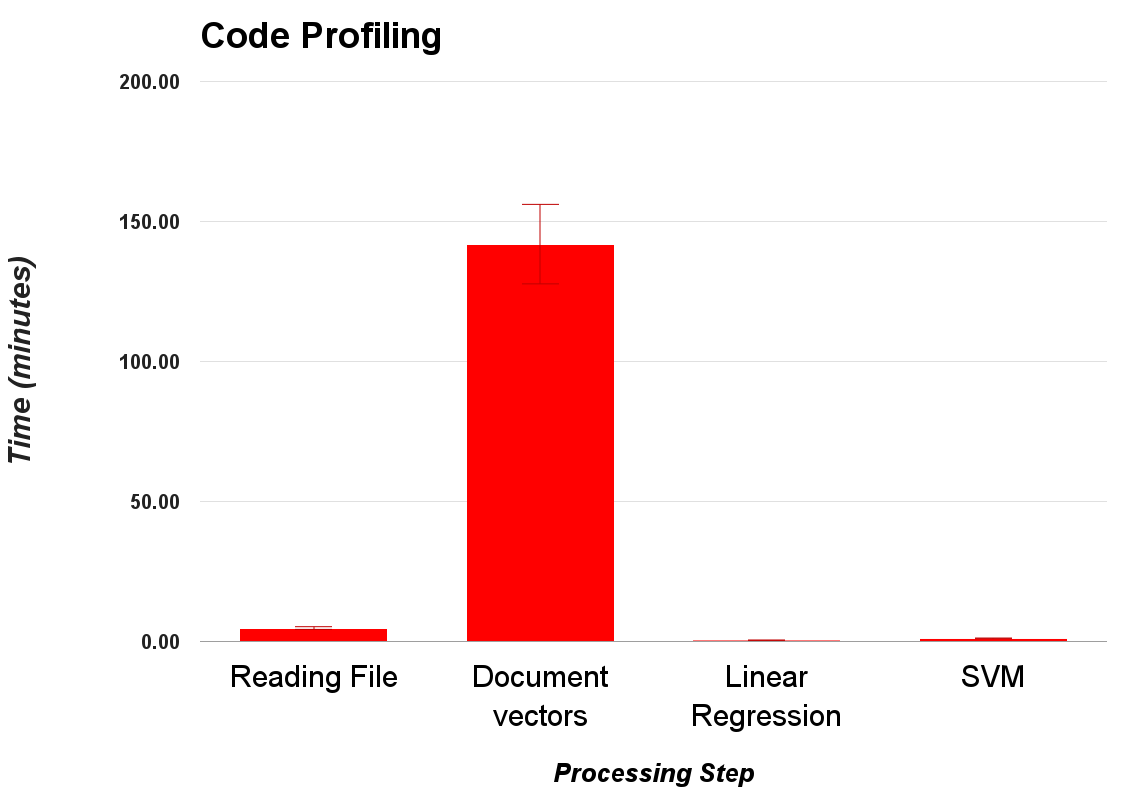}
    \caption{Code Profiling Statistics}
    \label{fig:code-profiling}
\end{figure*}

\section{Implementation}
    This section briefly discusses the implementation specifics chosen for the code-base, and the architectural decisions that govern this particular framework. 
    This is helpful to know in case a new application to be chosen to act as a pipeline into, or a sink of the data processed by the Rapid-Rate framework.

    The primary language of implementation of this research project is Python, and the other frameworks and components have been described in the Tech Stack section. (Section \ref{Tech stack})

    \subsection{Off-line learning module} \label{offline_module}
        The off-line learning module is comprised of the shallow neural-network implementation using the library Gensim (Section \ref{gensim}). 
        The algorithm of this module is shown in Algorithm \ref{offline_module_algo}.

    \subsection{On-line real-time prediction engine}
        The on-line prediction module is the real-time component submitted to Spark (Section \ref{spark}) workers. 
        Each of the spark workers listen to a distributed queue in Kafka (Section \ref{kafka}). 
        As and when the unlabeled documents are streamed, the engine reads the streamed data and predicts a score before posting it into OpenTSDB (Section \ref{opentsdb}). 
        The general algorithm for handling the streaming data is described in Algorithm \ref{online_module}. 
        A sub-section of this module, the algorithm used to infer sentiment, is described in Algorithm \ref{infer_sentiment}.

    \subsection{Tech stack} \label{Tech stack}
        The sub-section described the composition of the tech-stack used to implement the Rapid-Rate framework.

        \subsubsection{Python}
            Document processing \& cleaning, pipelining, real-time prediction\cite{python}.

        \subsubsection{Gensim} \label{gensim}
            Python library. 
            The Doc2Vec class in Gensim is used to learn document vector representations\cite{doc2vec_api}.
        
        \subsubsection{Scikit-Learn}
            Python library. 
            Enable the computation of linear regression and support vector regression models\cite{scikit_learn}.
        
        \subsubsection{Apache Kafka} \label{kafka}
            Distributed messaging queue. 
            Used to receive new unlabeled documents, and is also a potential sink for dumping the results of the result ratings of unlabeled documents\cite{kreps2011kafka}.
        
        \subsubsection{Apache Spark Streaming} \label{spark}
            The real-time prediction component written in Python is submitted as a Spark Streaming job\cite{zaharia2012discretized}.
        
        \subsubsection{OpenTSDB} \label{opentsdb}
            Time series database. 
            This database houses each individually predicted document rating and their corresponding time-stamps\cite{opentsdb}.

        \begin{figure*}[ht]
            \centering
            \includegraphics[width=\textwidth]{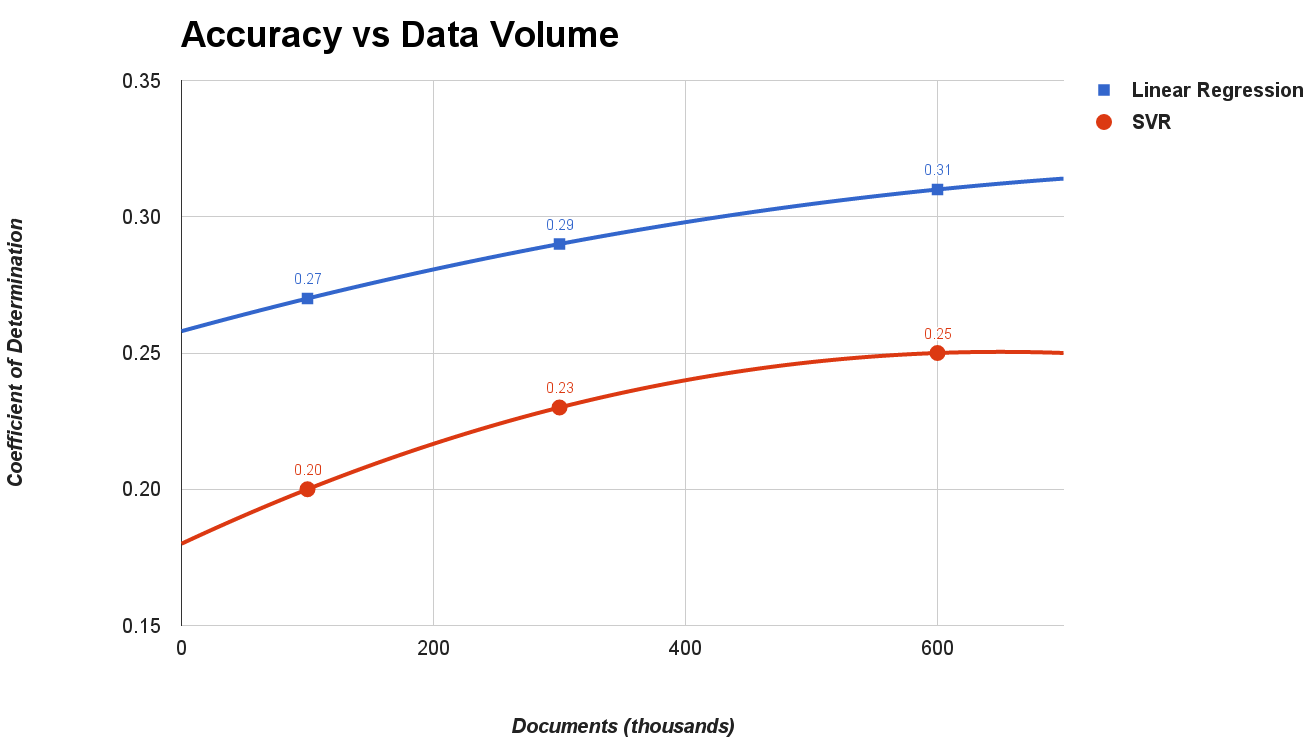}
            \caption{Coefficient of Determination vs. Training corpus volume}
            \label{fig:accuracy-v-data}
        \end{figure*}
        
        \subsubsection{Bosun}
            Bosun comprises of 2 things - a graphical interface with which to perform easy temporal scans, and a configurable alerting system based on the rules on the incoming values\cite{bosun_repo}. 

        The chosen tech-stack is state-of-the-art for framework design and development, with a strong developer community and guaranteed long-term support for each, making the framework future-proof for the foreseeable time.

\begin{figure*}[ht]
    \centering
    \includegraphics[width=0.9\textwidth]{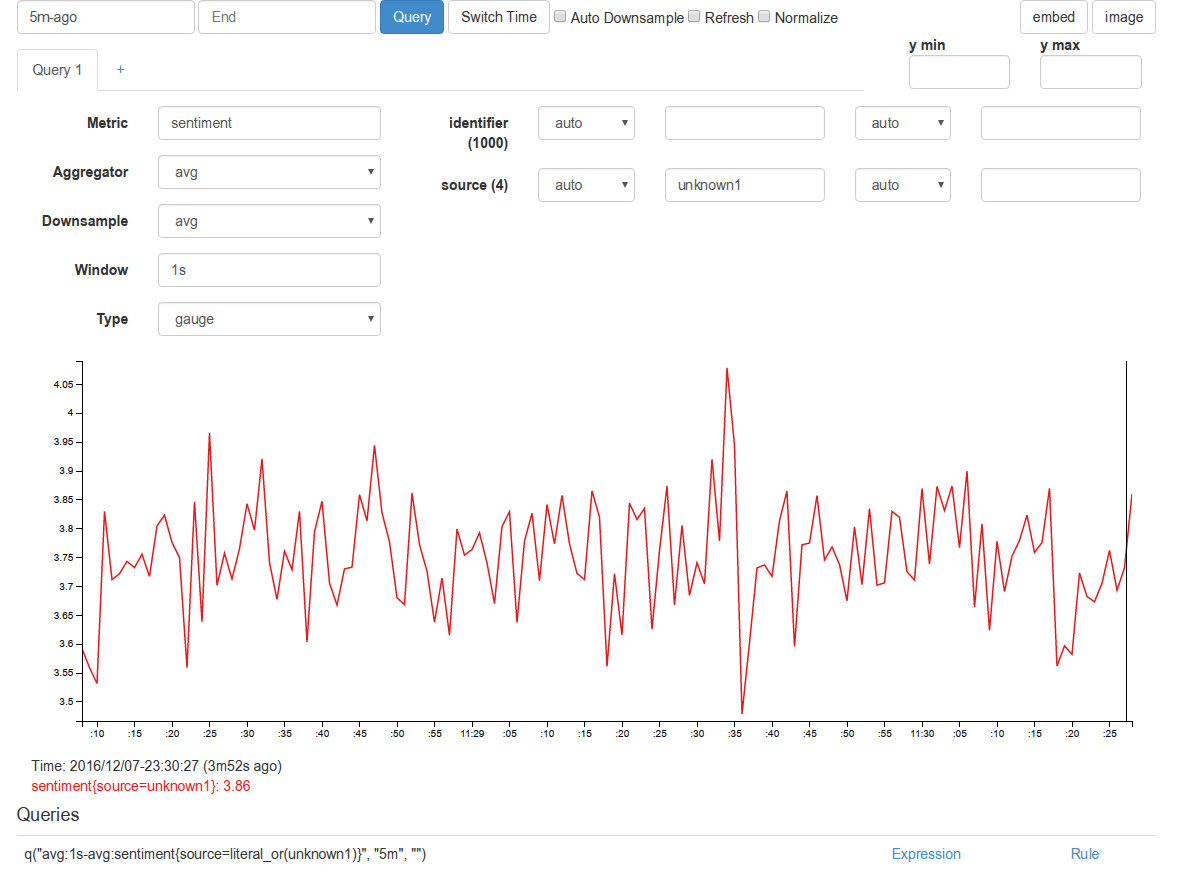}
    \caption{Time Series Analytics Dashboard}
    \label{fig:time-series-analytics-dashboard}
\end{figure*}

\vspace{5mm}

\section{Evaluation}

    \subsection{Testbed}
        The testbed is integrated into the actual framework, and the processors for model training evaluate the model accuracy.
        The tests were carried out on an Amazon document corpus of 600,000 unique product reviews.
        The complete dataset of Amazon reviews is openly available to download\cite{amazon_datasets}.

    \subsection{Metrics}
        The experimental results are being evaluated for goodness of model fit using the co-efficient of determination metric \cite{jaeger1990statistics}.
        The co-efficient of determination conveys the quantized amount of signal of the input vector that is being captured by the trained model by validating it against a test set. 
        The value of the co-efficient of determination is derived from the below equations:

        \begin{equation}
            \displaystyle SS_{tot} = \sum_{i} (y_i - \hat{y})^2
        \end{equation}

        \begin{equation}
            \displaystyle SS_{res} = \sum_{i} (f(x)_i - \hat{y})^2
        \end{equation}

        \begin{equation} \label{eq:r_squared}
            \displaystyle R^2 = 1 - \frac{SS_{res}}{SS_{tot}}
        \end{equation}

        The $R^2$ score derived in Equation \ref{eq:r_squared} is the the co-efficient of determination use to predict the goodness of fit for each of the machine learning models learned.

    \subsection{Experimental Results}
        The accuracy scores of the experimental results are presented in Table \ref{results}. The score presented in this table is the co-efficient of determination or $R^2$ score.

        \begin{table}[H] \caption{Evaluation results} \label{results}
            \centering
            \resizebox{0.4\textwidth}{!}{
                \begin{tabular}{| c | c |}
                    \hline
                    \textbf{Modeling Method} & \textbf{$R^2$ Score} \\
                    \hline
                    Linear Regression & 0.31 \\
                    \hline
                    Support Vector Regression & 0.25 \\
                    \hline
                \end{tabular}
            }
        \end{table}

        The positive scores of both the experiment analyses indicate that the model does capture a good amount of the sentiment in each of the labeled documents, and ensures a reasonable amount of accuracy for the prediction of the real-time unlabeled documents.

        As indicated in Figure \ref{fig:accuracy-v-data}, the current trend seems to suggest at least a polynomial-function improvement in the accuracy with respect to increasing the number of documents to be trained on. 
        The initial experimentation was done on a corpus of 100,000 labeled documents. 
        The subsequent experiments used corpora with 300,000 documents and eventually 600,000 documents.

        Also, from the code profiling information presenting in Figure \ref{fig:code-profiling}, it is evident that the unsupervised learning segment of obtaining the document vectors of the labeled documents is the longest running part of the off-line learning module. 
        In comparison, the document processing and statistical model training is negligible.

\vspace{5mm}

\section{Conclusions}
    To summarize, this research project encapsulated the below objectives:
    \begin{itemize}
        \item Evaluating the usage of document embeddings for a regression task
        \item Building a framework to predict real-time sentiment trends
    \end{itemize}

    Given that the attempt is to infer sentiment from unstructured data, the result of having a positive $R^2$ value is promising (Table \ref{results}).
    It indicates that the model captures a significant amount of the signal present in the independent variable.
    It can hence be relied to a degree of confidence to predict a close enough value to the actual intended document sentiment.

    Additionally, this work has resulted in a generic framework that can be adopted to build a sentiment analytics dashboard on a user friendly interface. 
    This framework can be extended to process any type of document content both for vector training and for the real-time sentiment prediction module.

    The overall framework has been built using state-of-the-art product offerings for neural-net document embedding libraries, distributed stream processing in Spark, distributed queue architecture in Kafka, and the stable machine learning framework which is a part of scikit-learn. 

    The author believes this framework can be used both in industrial settings as well as in evaluation testbeds for any document embedding and machine learning libraries used as an ensemble architecture. The hypothesis made prior to undertaking this research project, as described in the Problem Statement (Section \ref{problem_statment}) has been validated and proved to be true.

    The framework itself, is open-sourced, and is free to be used to replicate the results or to be extended independently\cite{rapid_rate}.

\vspace{5mm}

\section{Future Work} \label{future_work}
    With respect to the statistical models trained, the current approach supports only the simplest regression models i.e. Linear Regression and Support Vector Regression (SVR). 
    This could be expanded to offer additional training mechanisms that can integrate more sophisticated regression algorithms for a higher level of accuracy.
     
    Additional channels could be chosen as the culmination point of the newly rated unlabeled documents.
    For now, the data is stored in a time-series database.
    But this can easily be converted into something more generic, like publishing it to another Kafka queue.
    Multiple consumer applications can, thus, benefit from the stream of insights published to the stream.

    With reference to the natural language understanding challenge described in Section \ref{cross_domain_applicability}, a proposed solution could be to merely obfuscate the entities being spoken about the the document for which the sentiment analysis model is being trained for. 
    Similarly, the obfuscation also needs to be done for the test dataset. 
    Since this removes the dependence of the sentiment in the document to the sentiment expressed in it, it would logically allow for a more generic prediction model. 
    Although this won't completely mitigate the presence of domain-specific jargon, it does allow for a more accurate prediction model, and remains to be evaluated.

    Another potential aim for future work is to reduce the time taken for the initial document vector training of the off-line model.
    As shown in the code profiling statistics present in Figure \ref{fig:code-profiling}, it is evident that a significant reduction in the cost of training the initial model could be achieved if the document vectors training time were to be reduced. 
    This can be achieved by the horizontal scaling described in Section \ref{dist_docvec_learning}.

\vspace{5mm}

\section{Acknowledgments}
    The author would like to thank Dr. Paulo Alencar, for his guidance, feedback and suggestions during the conceptualization of this research project. 

    The author would also like to thank Dr. Olga Vechtomova for her feedback and ideas on document evaluation and sentiment polarity tasks.

\vspace{5mm}

\bibliographystyle{abbrv}
\balance
\bibliography{rapid-rate-report}

\begin{thebibliography}{10}

\bibitem{amazon_datasets}
Amazon product review data.
\newblock \url{http://jmcauley.ucsd.edu/data/amazon/}.
\newblock Accessed: 2016-12-12.

\bibitem{bosun_repo}
Bosun - an open-source, mit licensed, monitoring and alerting system by stack
  exchange.
\newblock \url{https://github.com/bosun-monitor/bosun}.
\newblock Accessed: 2016-12-13.

\bibitem{continuous_function}
Continuous function.
\newblock
  \url{https://www.encyclopediaofmath.org/index.php/Continuous_function}.
\newblock Accessed: 2016-12-09.

\bibitem{doc2vec_api}
Doc2vec (gensim) - deep learning via the distributed memory and distributed bag
  of words models using either hierarchical softmax or negative sampling.
\newblock \url{https://radimrehurek.com/gensim/models/doc2vec.html}.
\newblock Accessed: 2016-12-11.

\bibitem{opentsdb}
Opentsdb - the scalable time series database.
\newblock \url{https://github.com/OpenTSDB/opentsdb}.
\newblock Accessed: 2016-12-13.

\bibitem{python}
Python.
\newblock \url{https://www.python.org/}.
\newblock Accessed: 2016-12-13.

\bibitem{rapid_rate}
Rapid-rate, a framework for semi-supervised real-time sentiment trend detection
  in unstructured big data.
\newblock \url{https://github.com/v1n337/rapid-rate}.
\newblock Accessed: 2016-12-13.

\bibitem{regression_analysis}
Regression analysis.
\newblock
  \url{https://www.encyclopediaofmath.org/index.php/Regression_analysis}.
\newblock Accessed: 2016-12-11.

\bibitem{scikit_learn}
scikit-learn: Machine learning in python.
\newblock \url{http://scikit-learn.org/stable/}.
\newblock Accessed: 2016-12-13.

\bibitem{tensorflow_word2vec}
Vector representations of words.
\newblock
  \url{https://www.tensorflow.org/versions/r0.12/tutorials/word2vec/index.html}.
\newblock Accessed: 2016-12-08.

\bibitem{basak2007support}
D.~Basak, S.~Pal, and D.~C. Patranabis.
\newblock Support vector regression.
\newblock {\em Neural Information Processing-Letters and Reviews},
  11(10):203--224, 2007.

\bibitem{bespalov2011sentiment}
D.~Bespalov, B.~Bai, Y.~Qi, and A.~Shokoufandeh.
\newblock Sentiment classification based on supervised latent n-gram analysis.
\newblock In {\em Proceedings of the 20th ACM international conference on
  Information and knowledge management}, pages 375--382. ACM, 2011.

\bibitem{bottou2010large}
L.~Bottou.
\newblock Large-scale machine learning with stochastic gradient descent.
\newblock In {\em Proceedings of COMPSTAT'2010}, pages 177--186. Springer,
  2010.

\bibitem{buneman1996query}
P.~Buneman, S.~Davidson, G.~Hillebrand, and D.~Suciu.
\newblock A query language and optimization techniques for unstructured data.
\newblock In {\em ACM SIGMOD Record}, volume~25, pages 505--516. ACM, 1996.

\bibitem{cameron1997r}
A.~C. Cameron and F.~A. Windmeijer.
\newblock An r-squared measure of goodness of fit for some common nonlinear
  regression models.
\newblock {\em Journal of Econometrics}, 77(2):329--342, 1997.

\bibitem{drake2008sentiment}
A.~Drake, E.~K. Ringger, and D.~A. Ventura.
\newblock Sentiment regression: Using real-valued scores to summarize overall
  document sentiment.
\newblock 2008.

\bibitem{gonzalez2011identifying}
R.~Gonz{\'a}lez-Ib{\'a}nez, S.~Muresan, and N.~Wacholder.
\newblock Identifying sarcasm in twitter: a closer look.
\newblock In {\em Proceedings of the 49th Annual Meeting of the Association for
  Computational Linguistics: Human Language Technologies: short papers-Volume
  2}, pages 581--586. Association for Computational Linguistics, 2011.

\bibitem{jaeger1990statistics}
R.~M. Jaeger.
\newblock {\em Statistics: A spectator sport}, volume~5.
\newblock Sage, 1990.

\bibitem{kreps2011kafka}
J.~Kreps, N.~Narkhede, J.~Rao, et~al.
\newblock Kafka: A distributed messaging system for log processing.
\newblock In {\em Proceedings of the NetDB}, pages 1--7, 2011.

\bibitem{le2014distributed}
Q.~V. Le and T.~Mikolov.
\newblock Distributed representations of sentences and documents.
\newblock In {\em ICML}, volume~14, pages 1188--1196, 2014.

\bibitem{mikolov2013efficient}
T.~Mikolov, K.~Chen, G.~Corrado, and J.~Dean.
\newblock Efficient estimation of word representations in vector space.
\newblock {\em arXiv preprint arXiv:1301.3781}, 2013.

\bibitem{mikolov2013distributed}
T.~Mikolov, I.~Sutskever, K.~Chen, G.~S. Corrado, and J.~Dean.
\newblock Distributed representations of words and phrases and their
  compositionality.
\newblock In {\em Advances in neural information processing systems}, pages
  3111--3119, 2013.

\bibitem{pang2002thumbs}
B.~Pang, L.~Lee, and S.~Vaithyanathan.
\newblock Thumbs up?: sentiment classification using machine learning
  techniques.
\newblock In {\em Proceedings of the ACL-02 conference on Empirical methods in
  natural language processing-Volume 10}, pages 79--86. Association for
  Computational Linguistics, 2002.

\bibitem{su2015genetic}
B.-h. Su and Y.-l. Wang.
\newblock Genetic algorithm based feature selection and parameter optimization
  for support vector regression applied to semantic textual similarity.
\newblock {\em Journal of Shanghai Jiaotong University (Science)}, 20:143--148,
  2015.

\bibitem{weissman2016natural}
G.~E. Weissman, M.~O. Harhay, R.~M. Lugo, B.~D. Fuchs, S.~D. Halpern, and M.~E.
  Mikkelsen.
\newblock Natural language processing to assess documentation of features of
  critical illness in discharge documents of acute respiratory distress
  syndrome survivors.
\newblock {\em Annals of the American Thoracic Society}, 13(9):1538--1545,
  2016.

\bibitem{zaharia2012discretized}
M.~Zaharia, T.~Das, H.~Li, S.~Shenker, and I.~Stoica.
\newblock Discretized streams: an efficient and fault-tolerant model for stream
  processing on large clusters.
\newblock In {\em Presented as part of the}, 2012.

\end{thebibliography}

\end{document}